\def\year{2020}\relax
\documentclass[letterpaper]{article} 
\usepackage{aaai20}  
\usepackage{times}  
\usepackage{helvet} 
\usepackage{courier}  
\usepackage[hyphens]{url}  
\usepackage{graphicx} 
\usepackage{graphicx}  
\usepackage[ruled,vlined,linesnumbered]{algorithm2e} 

\usepackage{url}            
\usepackage{booktabs}       
\usepackage{amsfonts}       
\usepackage{microtype}      

\usepackage{times}
\usepackage{url}
\usepackage{enumitem}
\usepackage{graphicx}
\usepackage[cmex10]{amsmath}
\usepackage{amsthm,amssymb}
\usepackage{amsmath}
\usepackage{xspace}
\usepackage{tabularx}

\usepackage{float}

\usepackage{rotating}
\usepackage{tikz}

\usepackage{blkarray}
\usepackage{bbm}

\usepackage{graphicx,amsmath,amssymb,amsthm,xspace,microtype,soul,bbm}
\usepackage{bm,multirow,subfigure,makecell,booktabs,array}

\usepackage{mathtools,lipsum} 

\usepackage{dsfont}

\theoremstyle{definition}

\theoremstyle{definition}

\newtheorem{proposition}{Proposition}

\newtheorem{assumption}{Assumption} 
\newtheorem{example}{Example}

\newtheorem{remark}{Remark}

\usepackage{mathtools}  

\newtheorem{task}{Task}
\newcommand{\kl}{D}
\newcommand{\mi}{I}
\newcommand{\se}{H}
\newcommand{\E}{\mathbb{E}}

\newcommand{\dob}[1]{do(#1)}

\newcommand{\Obs}{Y}
\newcommand{\obs}{y}
\newcommand{\Out}{Z}
\newcommand{\out}{z}

\newcommand{\dec}{action-effect transfer learning task\xspace}
\newcommand{\Dec}{Action-effect transfer learning task\xspace}

\newcommand{\defi}{\emph}

\DeclareMathOperator*{\SumInt}{%
	\mathchoice%
	{\ooalign{$\sum$\cr\hidewidth$\displaystyle\int$\hidewidth\cr}}
	{\ooalign{\raisebox{.14\height}{\scalebox{.7}{$\textstyle\sum$}}\cr\hidewidth$\textstyle\int$\hidewidth\cr}}
	{\ooalign{\raisebox{.2\height}{\scalebox{.6}{$\scriptstyle\sum$}}\cr$\scriptstyle\int$\cr}}
	{\ooalign{\raisebox{.2\height}{\scalebox{.6}{$\scriptstyle\sum$}}\cr$\scriptstyle\int$\cr}}
}

\usepackage{tikz}
\usetikzlibrary{arrows,backgrounds,fadings}
\usetikzlibrary{shapes,plotmarks}
\usetikzlibrary{bayesnet}

\tikzstyle{var}=[fill=none,draw=none]
\tikzstyle{hid}=[circle,fill=none,draw=gray,text=black]

\usepackage{xr}
\newcommand{\keepifspace}{}

\title{Causal Transfer for Imitation Learning \\ and Decision Making under Sensor-shift}
\author{\Large \textbf{Jala Etesami, Philipp Geiger }\\ 
\textsuperscript{\rm }Bosch Center for Artificial Intelligence - BCAI \\ 
Robert Bosch GmbH\\
71272 Renningen, Germany\\
Jalal.Etesami@de.bosch.com,\\
Philipp.W.Geiger@de.bosch.com 
}

\begin{document}

\maketitle

\begin{abstract}
Learning from demonstrations (LfD) is an efficient paradigm to train AI agents. 
But major issues arise when there are differences between (a)  the demonstrator's own sensory input, (b) our sensors that observe the demonstrator and (c) the sensory input of the agent we train.

In this paper, we propose a causal model-based framework for transfer learning under such ``sensor-shifts'', for two common LfD tasks:
(1) inferring the effect of the demonstrator's actions and (2) imitation learning.
%
First we rigorously analyze, on the population-level, to what extent the relevant underlying mechanisms (the action effects and the demonstrator policy) can be identified and transferred from the available observations together with prior knowledge of sensor characteristics. And we device an algorithm to infer these mechanisms.
Then we introduce several proxy methods which are easier to calculate, estimate from finite data and interpret than the exact solutions, alongside theoretical bounds on their closeness to the exact ones.
We validate our two main methods on simulated and semi-real world data.
\end{abstract}

%

\section{Introduction}\label{sec:intro}

%

\paragraph{Motivation.}
Learning from demonstrations is an important paradigm to train AI agents \cite{argall2009survey,schaal1999imitation,ho2016generative,jeon2018bayesian}.
Ideally, one would like to harness as much \emph{cheaply available (and relevant) demonstrator data} as possible. 
But major issues arise when there are \emph{differences between the sensors} of demonstrator, us and agent we train.
When ignoring such issues, or addressing them in a naive way, wrong and potentially harmful conclusions can result: about demonstrator's behavior and the demonstrator's actions' effects on the environment.%

%
%

\begin{figure}[t]
	\centering
	\includegraphics[scale=.6]{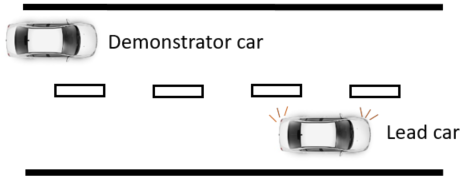}
	\caption{In highway drone data, the \emph{indicator light} of the lead car would be missing, introducing a \emph{hidden common cause} between acceleration of demonstrator car and lane changing behavior of the lead car. }\label{fig:example1}
\end{figure}

\begin{example}[Highway drone data]
\label{expl:i}
In the development of self-driving cars, recently drones have been deployed to fly over highways and record the behavior of human-driven cars \cite{highDdataset,zhan2019interaction}.
Clearly, in such drone recordings, some crucial variables are either \emph{more noisy} than 
observed from within the car, or completely \emph{missing}, such as \emph{indicator lights}.

Assume we want to use such data 
to learn, say, how an acceleration action $A$ of a ``demonstrator car'' affects the lane changing behavior $Z$ of a ``lead car'' in front of it on the slower lane, as depicted in Figure \ref{fig:example1}.
Slightly simplifying reality, assume the indicator light of the lead car serves as a perfect coordination device: it is on if and only if, subsequently, (1) the demonstrator car decelerates and (2) the lead car changes lane to the fast lane.
Now assume we just use the variables recorded in the drone data, where the indicator light is not contained, estimate $P(Z|A)$ from it, and naively consider it as the \emph{causal effect of $A$ on $Z$}. 

This leads us to the conclusion that an agent in the place of the demonstrator can arbitrarily chose any acceleration or deceleration action as $A$, and the lead car will perfectly adapt $Z$ and only change lane when agent decelerates -- which in practice can lead to crashes. In the language of causal models \cite{pearl2009causality,spirtes2000causation}, the indicator light is a \emph{hidden common cause (confounder)}.
\end{example}

\paragraph{Main tasks, approach and contributions:}
In this paper, we address learning from demonstrations (LfD) under \emph{sensor-shift}, i.e., when 
there are differences between (a)  the demonstrator's own sensory input, (b) our sensors that observe the demonstrator and (c) the sensory input of the agent we train. 
Specifically, we consider two closely related ``subtasks'' of LfD: (1) inferring the effect of the demonstrator's decisions (as in Example \ref{expl:i}) and (2) imitating the demonstrator.
%

Our approach is based on causal models \cite{pearl2009causality,spirtes2000causation,peters2017elements}, which allow us to generalize from data beyond i.i.d.\ settings.
The idea is that, while some modular causal mechanisms that govern the data vary (the sensors), other mechanisms are \emph{invariant} (e.g., the action-effect). 

Our main contributions are: 
\begin{itemize}

\item We rigorously analyze, on the population-level, 
to what extent the relevant underlying mechanisms (the action-effect and the demonstrator policy) can be \emph{identified} and transferred from the available observations together with prior knowledge of sensor characteristics (Sections \ref{sec:general}, \ref{sec:deex}, \ref{sec:deexlin} and \ref{sec:imex}). And we propose algorithms to calculate them (Algorithms \ref{alg:gen} and \ref{alg:lin}). 

\item We introduce several \emph{proxy methods} (Sections~\ref{sec:deav} and \ref{sec:imav}) which are easier to calculate, estimate from finite data and interpret than the exact solutions, alongside theoretical bounds on their closeness to the exact ones 
(Propositions~\ref{thm:weightedproxy}, \ref{pro:imi2} and \ref{pro:imi5}). (Proofs are in the supplement\footnote{The supplement can be found at ``\url{https://doi.org/10.5281/zenodo.3549981}''.} of this paper.) 
\item We conduct \emph{experiments} to validate our two main methods on \emph{simulated and semi-real world} highway drone data used for autonomous driving (Section \ref{sec:exp}). 

\end{itemize}

\section{Related work}

%
%
%
%
%
%
%

\emph{Learning from demonstrations} (LfD) \cite{argall2009survey} is a broad area, with two concrete tasks being the ones we also consider in this paper: (1) inferring the effect of action on outcome given observation (we call it ``action-effect'' in our a-temporal framework, while in the language of \cite{argall2009survey} this is called the ``system model'' or ``world dynamics''), and (2) imitation learning (see next paragraph).
Generally in LfD, the problem that sensors differ between demonstrator, observer and target AI agent has been considered \cite{argall2009survey,ude2004programming,atkeson1997robot}.
In the language of \cite{argall2009survey}, this is described as the ``recording mapping'' or ``embodiment mapping'' not being the identity. However, we are not aware of any treatment of this problem which is as \emph{systematic and general} as ours in terms of  guarantees on exact and approximate identifiability. Instead, approaches are practically-focused, tailored to specific, say, robot tasks \cite{ude2004programming,atkeson1997robot}.




Within LfD, \emph{imitation learning} means learning to perform a task from expert demonstrations \cite{ho2016generative,muller2006off}. 
There are two main approaches to address this problem: behavioral cloning \cite{pomerleau1991efficient}, which we are focusing on, 
and inverse reinforcement learning (IRL) \cite{ng2000algorithms,ziebart2008maximum}. 


The problem of bounding as well as transferring and integrating \emph{causal relations} across different domains has been studied by \cite{balke1994counterfactual,bareinboim2014generalizability,magliacane2017causal}. 
But all this work does not consider the training of AI agents.
Within causal modelling, maybe closest related to our paper are \cite{bareinboim2015bandits,forney2017counterfactual,zhang2017transfer,geiger2016experimental}, who also study the integration of data from heterogeneous settings for training agents (often with latent confounders and from a multi-armed bandit perspective).

%
For example, \cite{zhang2017transfer} tackle the problem of transferring knowledge across bandit agents in settings where causal effects cannot be identified by standard learning techniques. Their approach consists of two steps: (1) deriving bounds over the effects of selecting arms and (2) incorporating these bounds to search for more promising actions.
However, when bounding the causal effect, they focus on binary variables, while we consider arbitrary finite as well as continuous ranges (which are highly relevant in practice) and they do not focus on general sensor-shifts.

The authors of \cite{causalconfusion} study ``causal confusion" in causal-model-free imitation learning. There, additional observations can lead to worse performance due to the mechanism (policy) that generates them differing between demonstrations and target environment. 
However, in their model they assume that both the demonstrator and the imitator have (at least) the same observations.
This is not always the case, and therefore our treatment allows the observations to differ.

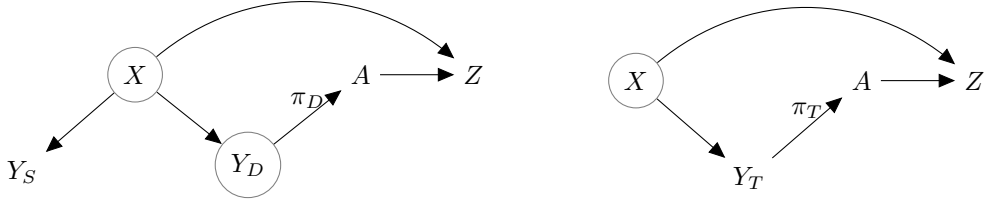
\begin{figure*}[t]
	\centering 
	\begin{tikzpicture}[scale=1]
	\node at (3, 0) (A) {$A$};
	\node[var] at (4.5, 0) (R) {$\Out$};
	\node[hid] at (0, 0) (X) {$X$};
	\node[var] at (-1.5, -1.3) (Ws) {$\Obs_S$};
	\node[hid] at (1.5, -1.2) (Wd) {$\Obs_D$};
	
	\draw[->] (A) to (R);
	\draw[->] (X) to (Wd);
	\draw[->] (X) to (Ws);
	\draw[->] (Wd) to node [above,midway] {$\pi_D$} (A);
	\draw[->,bend left=40]  (X) to (R);
	\end{tikzpicture}
	\begin{tikzpicture}[scale=1]
	\node at (3, 0) (A) {$A$};
	\node[var] at (4.5, 0) (R) {$\Out$};
	\node[hid] at (0, 0) (X) {$X$};
	\node[var] at (1.5, -1.3) (Wt) {$\Obs_T$};
	
	\draw[->] (A) to (R);
	\draw[->] (X) to (Wt);
	\draw[->] (Wt) to node [above,midway] {$\pi_T$} (A);
	\draw[->,bend left=40]  (X) to (R);

	\node[var] at (-1.5, 0) (Ws) {$\phantom{\Obs_S}$};
	\end{tikzpicture}
	\caption{Causal DAGs. \textbf{Left:} source domain. \textbf{Right:} target domain. Circle means hidden to us.} 
	\label{fig:cdag}
\end{figure*}

\section{Background}
\label{sec:background}

\paragraph{Conventions:}
We use $D(\cdot||\cdot)$, $H(\cdot)$, and $I(\cdot;\cdot|\cdot)$ to denote the Kullback-Leibler (KL) divergence, entropy, and mutual information, respectively \cite{cover2012elements}. 
We consider both, discrete and continuous random variables; $\SumInt$ stands for the sum or integral, accordingly; $P(W)$ for the distribution of a variable $W$, and $p(w)$ for the density at value $W=w$. 
If not stated otherwise, we assume that distributions have full support\footnote{Full support is a commonly made \cite{pearl2009causality} but non-trivial assumption, important for identifiability.} and densities.

\newcommand{\PA}{\mathit{PA}}
\paragraph{Causal models:}
According to Pearl's definition \cite{pearl2009causality}, a \defi{causal model} is an ordered triple $(U, V, E)$, where $U$ denotes a set of \emph{exogenous variables} whose values are determined by factors outside the model (not observable); $V$ is a set of \emph{endogenous variables} whose values are determined within the model; and $E$ is a set of \emph{structural equations} that express, for each endogenous variable $W \in V$, the \emph{mechanism} of how $W$ is generated by certain other endogenous and exogenous variables. Namely, for all $W\in V$, we have 
$$
W = f_W(\PA_W, U_W),
$$
 where $f_W(\cdot,\cdot)$ is a function and $\PA_W$ denotes the \defi{parent set} of variable $W$. $W$ is called a \emph{child} of $\PA_W$.
This induces a joint distribution over the endogenous variables, which can be factorized as follows:
$$
P(V) = \prod_{W \in V} P(W|\PA_W).
$$
This factorization is usually expressed using a \defi{directed acyclic graph (DAG)}, in which nodes represent the endogenous variables and arrows are from parents to their children.
It is also possible that a sub-set of $V$ is hidden. In this case, we denote the hidden variable with circles in the DAG.

The \defi{post-interventional distribution} is defined by replacing a subset of structural equations without generating cycles in the DAG \cite{pearl2009causality}. More specifically, the post-intervention distribution after (\textit{atomic}) intervening on variable $W$ is defined by replacing $f_W(\PA_W, U_W)$ with value $w$ and it is denoted by
$
P\big(V| do(W = w)\big).
$

\section{Setting and problem formulation}

\subsection{General model of our setting}
\label{sec:model}

\paragraph{Causal models of source and target domain.} There are two domains, the \emph{source domain} where the \emph{demonstrator (agent)} observes and acts, and the \emph{target domain} where the \emph{target agent}, which we design, observes and acts.
(By domain we mean the complete causal model of environment, sensors, and agent.)
The two domains, including what is hidden and what is observed by us, are depicted by the two causal DAGs in Figure \ref{fig:cdag} over the following variables:
$X$ is the \emph{state} of the system, 
$A$ is the \emph{action} of the agent,
$Z$ stands for the \emph{outcome} (an abstract variable that could be, as in Example \ref{expl:i}, the state of cars in the the next time instance). 
Regarding \emph{observations}, we assume that in the source domain we have
$Y_D$, the \emph{demonstrator's input}, generated by the demonstrator's sensors,
$Y_S$, the \emph{spectator's -- i.e., our -- observation} of the state of the source system,
and in the target domain we have
$Y_T$, the \emph{input to the target agent} from the target agent's sensors.
We often denote distributions over variables (e.g. $P(Z)$) in the source and target domain by subscript $S$ and $T$, respectively (e.g., $P_S(Z)$ and $P_T(Z)$).
Let $\pi_D(A|Y_D)$ denote the \emph{policy of the demonstrator},
and  $\pi_T(A|Y_T)$ denote the \emph{policy of the target agent}.

\paragraph{Relationship between source and target domain, and what is known to us.} 
We assume that the two domains are related by sharing the same invariant mechanism for outcome given state and action, i.e., 
\begin{align*}
P_T(\Out|A, X)=P_S(\Out|A, X),
\end{align*}
so that we can drop the subscript and just write $P(\Out|A,X)$.
%
We assume we are given $P_S(\Out, A, \Obs_S)$ (or a sample of it), as well as the sensor characteristics\footnote{This may be based on performing an experimental system identification of the sensors or using physical knowledge.} $P_S(Y_S|X)$ and $P_T(Y_T|X)$.

\subsection{Problem formulation}
\label{sec:problem}

The overarching goal is to design the target agent that observes and successfully acts in the target domain, based on what we know from the source domain and its relation to the target domain.
We consider two specific tasks that serve this overarching goal:
\begin{task}[\Dec]
	\label{task:d}
	Infer $P_T(\Out|\dob{A}, Y_T)$, the effect of action $A$ on outcome $Z$ conditional on observation $Y_T$ in the target domain.%
	\footnote{Once the effect $P_T(\Out|Y_T, \dob{A})$ is inferred, what remains to be done for designing the target agent is to fix a utility function $u(\Out)$ on the outcome, and then pick the optimal $a$ by, say, maximizing $\E_T(u(\Out)|\dob{a}, y_T)$ w.r.t.\ $a$.}
\end{task}
\begin{task}[Imitation transfer learning task]
\label{task:i}
Learn a policy $\pi_T(A|\Obs_T)$ for the target agent (also called \defi{imitator} in this task) 
such that it behaves as similarly as possible to the demonstrator 
(details follow). 
\end{task}

\section{Basic step addressing both tasks: equations and algorithm}
\label{sec:general}

In this section, we make general derivations about our model (Section \ref{sec:model}), which serve as steps towards \emph{both}, the imitation and the action-effect transfer learning tasks.



\paragraph{Basic equation:}
Our model (Section \ref{sec:model}) implies the following equations, for all $\out, a, \obs$:

\begin{align}
&p_S(\out, a, \obs_S)  = \SumInt_x p_S(\obs_S|x) p_S(z, a, x)  \label{eqn:gen1} \\
&= \SumInt_{x, y_D} p_S(\obs_S|x) p(\out | a, x) \pi_D(a|\obs_D) p_S(\obs_D, x). \label{eqn:gen}
\end{align}
%
These are the basic equations that relates what is known -- $p_S(\out, a, \obs_S)$ (l.h.s. of~\eqref{eqn:gen1}) -- to what we would like to know (r.h.s.\ of~\eqref{eqn:gen}): $\pi_D(a|\obs_D)$ for Task \ref{task:i} and $p(\out | a, x)$ for Task \ref{task:d}.
More specifically, these equations \emph{constrain} the unknown quantities to a set of possibilities.
This is exactly the set up to which we can \emph{identify} \cite{pearl2009causality} them. 


\newcommand{\w}{P(\out, a, X)}
\newcommand{\F}{[P( y^i | x^j)]_{i,j=1}^{m, \ell}}
\newcommand{\vv}{P(\out, a, Y_S)}

\paragraph{Finite linear equation system in discrete case:}
Solving 
~\eqref{eqn:gen1} for $p_S(z, a, x)$ is an important intermediate step to addresses Task~\ref{task:d} and \ref{task:i} simultaneously, since $p_S(z, a, x)$ contains all the information that $p_S(z, a, \obs_S)$ contains about $\pi_D(a|y_D)$ and $p(\out | a, y_T)$.
(In particular, in the classical case of $Y_S=Y_T=Y_D=X$, $p_S(z, a, x)$ uniquely determines the latter two quantities via marginalization/conditioning.)
So let us for a moment focus on~\eqref{eqn:gen1}.
In the discrete case, it can be rewritten as the following collection of matrix equations.
Let  $\{ x^1, \ldots, x^\ell \}$ and $\{ y^1, \ldots, y^m \}$ be the range of $X$ and $\Obs_S$, respectively. 
Then, for all $ \out, a$,
	\begin{align}
	\underbrace{\left[ \!\!\!\begin{array}{c} P(\out, a, \obs^1) \\ \vdots \\ P(\out, a, \obs^m) \end{array} \!\!\right]}_{\vv \in \mathbb{R}^{m}} \!\!\!=\!\!\!
	\underbrace{\left[ \!\!\!\begin{array}{cc} P( y^1 | x^1) \cdots\!\!\!\!\!\! \!\!  & P( y^1 | x^\ell) \\ \vdots & \vdots \\   P( y^m | x^1)  \cdots\!\!\!\!\!\! \!\!  & P( y^m | x^\ell) \end{array}\!\!\! \right]}_{\F \in \mathbb{R}^{  m \times \ell} } \!
	\underbrace{\left[ \!\!\!\begin{array}{c} P(\out, a, x^1) \\ \vdots \\ P(\out, a, x^\ell) \end{array} \!\!\!\right]}_{P(\out, a, X)  \in \mathbb{R}^{\ell}} \!\!. \label{eqn:possible}
	\end{align}

\paragraph{Algorithm for solution set in discrete case:}
Algorithm \ref{alg:gen} yields a parametrization of the set of all possible solutions $P(\out, a, X) \in \mathbb{R}^\ell$ to~\eqref{eqn:possible}, for any $\out, a$. Specifically, it outputs the finite set of \emph{corner vectors} whose \emph{convex combinations parametrize} \emph{the solution set}.

It uses singular-value decomposition (SVD) to cope with non-invertibility, and then a routine inspired by the simplex algorithm to account for the constraint that the output has to be a proper probability distributions.\footnote{Since the left hand side of \eqref{eqn:possible} is a probability vector, it is not necessary to bound $P(z, a, x^i)$ by one.}

For the algorithm, w.l.o.g., we assume $m \leq \ell$ and that $\F$ has full rank (otherwise one removes linearly dependent rows).
Note that if $m=\ell$ and $\F$ is non-singular, then 
\eqref{eqn:possible} determines $\w$ uniquely, via a simple matrix inversion. 
Therefore, for this algorithm, the interesting scenario is 
	$m<\ell$. This is the case, e.g., in Example \ref{expl:i} -- the highway drone data where indicator lights are not recorded. 
\begin{algorithm}[t]
	\label{alg:gen}
	\caption{Finding solution set for \eqref{eqn:possible}}\label{alg:svd}
	\KwIn{$\vv$ (l.h.s.\ of \eqref{eqn:possible}), $\F$} 
	\KwOut{$\zeta_1, \ldots, \zeta_k \in \mathbb{R}^{\ell}$, such that their convex hull is the solution set to  \eqref{eqn:possible}
	}
	Rearrange columns of $\F$ such that $\F=[D\ E]$ and $D \in \mathbb{R}^{m \times m}$ is non-singular\;
	$U\Sigma V^T\leftarrow$ SVD of $\F$   \;
	\For{$i=1$ \KwTo $\ell-m$}{$e_i\leftarrow$ zero vector of length $\ell-m$ whose $i$th entry is one\;}
	$M\leftarrow \small{V\begin{bmatrix} \textbf{0} & \!\!\!\!\!\cdots\!\!\!\!\! & \textbf{0}  \\
    e_1 & \!\!\!\!\!\cdots\!\!\!\!\! & e_{\ell-m} \end{bmatrix}}$, 
	$b \leftarrow \begin{bmatrix}   D^{-1} \vv \\     \textbf{0}
\end{bmatrix}$\; 
 $i \leftarrow 1$\;
	\For{any sub-matrix $R$ of $M$ with dimension $(\ell-m)\times(\ell-m)$}{$\hat{b}\leftarrow$  
			the sub-vector of $b$ of length $\ell-m$ that corresponds to the selected rows of $M$\; 
	\If{$R^{-1}$ exists and $- M R^{-1}\hat{b}+b \geq 0$}{$\zeta_i\leftarrow -  M R^{-1}\hat{b}+b$\; $i \leftarrow i + 1$\;}
}
\end{algorithm}



\section{Approach to the \dec}
\label{sec:dea}

Let us now address Task~\ref{task:d} -- inferring the target domain's action-effect $P_T(\Out|\dob{A}, Y_T)$.

\begin{example}
	To illustrate what can go wrong when naively addressing this task, let us get back to the highway drone data (Example \ref{expl:i}). There, in the source domain, the indicator light is not observed by us, and for simplicity we assumed that there are no other variables, i.e., $Y_S$ is empty/constant. Our informal argument in that example can now be stated formally based on causal models (Section \ref{sec:background}):
	Observe that in the causal DAG (Figure \ref{fig:cdag}), $X, Y_D$ are \emph{hidden confounders} that introduce ``spurious correlations'' between $A$ and $Z$. Therefore, in the generic case, the naive guess $P_S(\Out|a)$ does not coincide with the actual action-effect $P_S(\Out|\dob{A})$ ($= P_T(\Out|\dob{A})$).
\end{example}


\begin{assumption}
	\label{asm:d}
	In this section, we assume the target agent observes the full state, i.e., $Y_T{=}X$.\footnote{Observability of $X$, similar as in Markov decision processes (MDPs), seems to be a good approximation to many real-world situations while at the same time keeping the analysis instructive. We make no assumption w.r.t.\ $Y_D$.}
\end{assumption}

Under Assumption \ref{asm:d}, we have 
$$
P_T(\Out|\dob{A}, Y_T) = P_T(\Out|\dob{A}, X) = P(\Out|A, X).
$$
 So Task~\ref{task:d} means inferring $P(\Out|A, X)$  (which could also be referred to as the (target domain's) ``dynamics'').
We now propose three methods, which differ w.r.t.\ the setting in which they are applicable and/or w.r.t.\ yielding exact or approximate solutions.

\subsection{Exact solution set in the discrete case}
\label{sec:deex}

In the case of all variables being discrete, we can build on our basic step in Section \ref{sec:general} to analytically find the set of possible action-effects $P(\Out|X, A)$ as follows: first we deploy Algorithm \ref{alg:gen} to get all possible $P(\Out,X, A)$, and then from this (simply by dividing by the marginals), we get  $P(\Out|X, A)$.

\subsection{Exact solution in the linear invertible continuous case}
\label{sec:deexlin}

In the continuous case, the general identification analysis -- the analysis of the solution set of \eqref{eqn:gen} -- is very difficult because the vectors space is infinite-dimensional.
Therefore let us here consider the special case of \emph{linear} relationships.


\newcommand{\Oa}{W}
\newcommand{\oa}{w}
\newcommand{\om}{F}
\newcommand{\No}{N}
\newcommand{\no}{n}
\newcommand{\Noz}{O}
\newcommand{\noz}{o}

\newcommand{\stax}{\left[\begin{smallmatrix}A \\ X\end{smallmatrix}\right]}

\newcommand{\samplelen}{\ell}

\newcommand{\idm}{\mathbf{1}}

\begin{assumption}
	\label{eqn:contas}
	In this Section \ref{sec:deexlin}, assume all relationships are linear, in particular, for matrices $D, E, F$,
	%
	\begin{align}
	\Obs_S &= \om X + \No ,\quad \label{eqn:c1} \\ 
	\Out &= [D\ E] \left[
	\begin{array}{c} 
	A \\ X
	\end{array} 
	\right]
	+ \Noz \label{eqn:c2}
	\end{align}
	with $\No, \Noz$ the usual noise terms that are independent of all other (non-descendant) variables.
\end{assumption}%
We propose Algorithm \ref{alg:lin} as (sample-level) method in this setting.

\begin{algorithm}[t]
	\label{alg:lin}
	\caption{Exact linear action-effect transfer method (sample-level)}
	\KwIn{sample $(\out_1, a_1, \obs_1), \ldots, (\out_\samplelen, a_\samplelen, \obs_\samplelen)$ from $P(\Out, A, Y_S)$; prior knowledge $F$, $\Sigma_{N N}$ (see \eqref{eqn:c1}); regularization parameter $\lambda$} 
	\KwOut{Estimates $\hat{D}, \hat{E}$ for the regression matrices $D, E$ (see \eqref{eqn:c2})}
	Calculate the empirical covariance matrices $\hat{\Sigma}_{Z A}, \hat{\Sigma}_{Z Y_S}, \hat{\Sigma}_{A Y_S}, \hat{\Sigma}_{Y_S Y_S}$ from the sample \label{line:s}\\
	Add a regularization term $\lambda \idm$ to $\hat{\Sigma}_{A A}$ and $\hat{\Sigma}_{Y_S Y_S}$\\
	Calculate the Schur complements\newline
	$S_1 := \hat{\Sigma}_{A A} - \hat{\Sigma}_{A Y_S} (\Sigma_{Y_S Y_S} - \Sigma_{\No \No })^{-1} \hat{\Sigma}_{Y_S A}$, 
	$S_2 := \hat{\Sigma}_{Y_S Y_S} - \Sigma_{\No \No } - \hat{\Sigma}_{Y_S A} \hat{\Sigma}_{A A}^{-1} \hat{\Sigma}_{A Y_S}$.\\
	Calculate the estimates 
	$\hat{D} := \hat{\Sigma}_{\Out A}  S_1^{-1} - \hat{\Sigma}_{\Out Y_S} (\hat{\Sigma}_{Y_S Y_S} - \Sigma_{\No \No })^{-1} \hat{\Sigma}_{Y_S A} S_1^{-1}$, and
	$\hat{E} := - \hat{\Sigma}_{\Out A} S_1^{-1} \hat{\Sigma}_{A Y_S} (\hat{\Sigma}_{Y_S Y_S} - \Sigma_{\No \No })^{-1} + \hat{\Sigma}_{\Out Y_S} S_2^{-1} ) F$
\end{algorithm}
\begin{proposition} 
	\label{thm:lin}
	Assume all variables have mean zero (otherwise center them).
	Furthermore, assume that $X$ and $Y_S$ have the same dimension, and that $F$ (in \eqref{eqn:c1}) is invertible.
	Then Algorithm~\ref{alg:lin} is \emph{sound} in the following sense: when replacing the empirical covariance matrices 
	$$
	\hat{\Sigma}_{Z A}, \hat{\Sigma}_{Z Y_S}, \hat{\Sigma}_{A Y_S}, \hat{\Sigma}_{Y_S Y_S}
	$$
	 in Line~\ref{line:s} by their population-level counterparts, and setting the regularization term $\lambda=0$, the output will be the true $D, E$ (in \eqref{eqn:c2}).
\end{proposition}%

\subsection{Average-based action-effect proxy in the general case}
\label{sec:deav}


The \emph{exact} general solution can be difficult to handle in terms of computation, estimation and analysis, and the linear case (Section \ref{sec:deexlin}) is of course restrictive.
Let us define the following \emph{average-based action-effect proxy} of the density $p(\out|x, a)$, for all $\out,x,a$, defined only based on things we do know (from the source domain):

\begin{align}
\tilde{p}(\out|x, a)\! :=\! \SumInt_{\obs_S} p_S(\out|\obs_S, a) p(y_S|x), \label{eqn:avp}
\end{align}

and let  $\tilde{P}(\Out|X, A)$ be the corresponding distribution.
The deviation between the average-based proxy and the ground truth it approximates can be bounded as follows: 

\begin{proposition}
	\label{thm:weightedproxy}
	We have%
	\footnote{In fact we bound the KL divergence between proxy and $p(\Out|X, A)$, but the expectation over $X, A$ is w.r.t.\ the \emph{source} domain, and therefore we have to write $p_S(\Out|X, A)$ on the l.h.s. of $\kl(\cdot \| \cdot)$. See also the proof.}
\[
\kl(P_S(\Out|X, A) \| \tilde{P}(\Out|X, A))
\leq \mi_S(X ; \Out | A, \Obs_S).
\]
\keepifspace{In particular, if $\Obs_S = f_{\Obs_S}(X)$ with $f_{\Obs_S}$ injective, then $\tilde{P}(\Out|X, A) = P(\Out|X, A)$.}
	Note that in the discrete case, the r.h.s.\ in turn can be bounded by an expression that is solely based on quantities, which we assumed to know:
	$
	\max_{P'(X)} \se_{X \sim P'(X)} (X | \Obs_S).
	$
\end{proposition}
%
%
%


\section{Approach to the imitation learning task }
\label{sec:imitask}

In this section, we address Task \ref{task:i}. To do so, we propose an imitator (the target agent) that selects a policy $\pi_T(A|\Obs_T)$ such that its behavior\footnote{Our notion of behavior is the conditional distribution of the action-outcome pair given the observation.} is as close as possible to the demonstrator.

Recall that, for the design of the imitator, what is available about the demonstrator is (a sample from) $P_S(\Obs_S, \Out, A)$.
However, the challenge is that the observation set of the demonstrator and the imitator may not be the same.
Therefore, we propose an imitator that behaves as close as possible to the demonstrator in case of perfect observation, i.e.,
 
\begin{align}\label{eq:imit_prob}
&\arg\min_{\pi_T}D\Big(P_T(A, \Out|X)||P_S(A, \Out|X)\Big).
\end{align}

It is worth noting that the imitator can also introduce additional constraints to this optimization problem according to its environment. 
Next, we give a simple example to illustrate what can go wrong when naively addressing the imitation task under sensor-shift. 
Then we propose methods for the problem in \eqref{eq:imit_prob} for several settings.

\begin{example}
Let us come back to Example \ref{expl:i} and Figure \ref{fig:example1}, where the indicator light perfectly correlates deceleration and lane changing. Let us add some modifications:
%
Assume we have the same sensors to observe the demonstrator as we have on board of the imitator's car, i.e., spectator's and imitator's sensors coincide, $P(Y_T|X) = P(Y_S|X)$.
And assume these sensors (similar to the drone) are missing the indicator light of the lead car (unlike the demonstrator's observation $Y_D$). 
Now, for the imitation task at hands, assume we naively take $\pi_T(a|\obs_T) := p_S(a | \Obs_S = \obs_T )$ as the imitator's policy.

This means that the imitator will accelerate and decelerate randomly, instead of, as the demonstrator, perfectly adapting these actions to the indicator light of the lead car (the indicator light is the actual source of variation in $A$ given $Y_D$, but the imitator just takes $P_S(A|Y_S)$ for a randomized policy). This will necessarily lead to crashes in the target domain -- whenever the lead car indicates and the imitator randomly decides to accelerate.
This issue can also be seen formally, based on the causal DAG (Figure \ref{fig:cdag}): there is a back-door path \cite{pearl2009causality} between action $A$ and outcome $Z$ that is not blocked by $Y_S$, and therefore, in the generic case, $P_S(\Out| \dob{A}, Y_S) \neq P_S(\Out| A, Y_S)$.
\end{example}

\subsection{Exact solution set in the discrete case}
\label{sec:imex}

\begin{assumption}\label{ass:im1}
Here we assume that both the demonstrator and the imitator have the same sensors\footnote{However, we relax this assumption in the next section.}, i.e., 
$$
P_S(\Obs_D|X) = P_T(\Obs_T|X).
$$
\end{assumption}
\begin{proposition}\label{pro:imi1}
Given Assumption \ref{ass:im1}, the solution of \eqref{eq:imit_prob} is 
$$
\pi_T(a|\Obs_T=\obs) := \pi_D(a|\Obs_D=\obs).
$$
\end{proposition}
Although this result introduces the optimal policy for the imitator, it is practical only if the imitator can infer $\pi_D(a|\Obs_D=\obs)$ using its observation from the source domain. 
In case of all variables being discrete, the imitator is able to do so using a set of finite linear equations similar to Section \ref{sec:general}. More precisely, 
\eqref{eqn:gen} leads to
\begin{align}\label{eq:cons}
\!\! P_S(a, \obs_S)\! =\!\! \sum_\obs  P_S(\obs_S|\Obs_D=\obs)P_S(a,\Obs_D=\obs).
\end{align}
\begin{assumption}
For the rest of Section \ref{sec:imitask}, we assume that $P_S(A, \Obs_S), P_S(\Obs_S|\Obs_D)$ are known to the imitator.
\end{assumption}
This forms a set of equation similar to \eqref{eqn:possible}. Algorithm \ref{alg:svd} (with input $P(a, Y_S)$, $[P( y_S^i | y_D^j)]_{i,j=1}^{m, \ell'}$, with $\ell'$ denoting the size of the range of $Y_D$) obtains the set of possible $P_S(a,\Obs_D)$ and consequently 
$$
\pi_D(a|\Obs_D)= \frac{P_S(a,\Obs_D)}{\sum_{a'}P_S(a',\Obs_D)}.
$$ 
\begin{remark}
Generally, it is important to mention that such assumptions can be weakened. But it will significantly increase the complexity of the problem by  essentially adding another layer of non-unique-identifiability of the joint from the conditional, e.g., $P_S(X, Y_S)$ from $P_S(Y_S|X)$. 
\end{remark}

\subsection{Average-based proxy in the general case}
\label{sec:imav}

Here, we propose proxy methods, which have the advantage that they can also be applied to the continuous case and may be easier to estimate/compute.
We do so for three different cases of sensor-shift. 

\paragraph{First case:} 
In this case, the imitator and the demonstrator have the same sensors in their domains, but the other sensors can be different, i.e., $P_T(\Obs_T|X)=P_S(\Obs_D|X)$.
Based on  Proposition \ref{pro:imi1}, the optimal policy for the imitator is indeed $\pi_D$. Thus, we propose the following policy: $\tilde{\pi}^{(1)}_T(a | Y_T=y):=\tilde{\pi}_D(a|\Obs_D=y)$, where the latter is defined by 

\begin{align}\label{eq:proxy_2}
&\SumInt_{\obs'} p_S(a|\Obs_S=y')p_S(\Obs_S=y'|\Obs_D=y).
\end{align}

\begin{proposition}\label{pro:imi2}
We have
$$
D( {\pi}_D|| \tilde{\pi}^{(1)}_T)\leq I_S(A;\Obs_D|\Obs_S).
$$ 
In the discrete case, additionally, the r.h.s.\ can be bounded by 
$$
I_S(A;\Obs_D|\Obs_S)\leq H(\Obs_D|\Obs_S).
$$
\end{proposition}

The above result implies that the proposed proxy and the demonstrator's policy are the same, when there exist deterministic relationship between the observation sets.
Next result goes beyond the policies and looks at the overall behavior of the system induced by this policy.

\begin{proposition}\label{pro:imi5}
The proposed proxy in \eqref{eq:proxy_2} implies that the KL-divergence in \eqref{eq:imit_prob} is bounded by 
$\begin{small}
D(\tilde{\pi}^{(1)}_T||\pi_D)
\end{small}$.
\end{proposition}

\paragraph{Second case:}
In this case, the spectator and the demonstrator have the same set of sensors in the source domain, i.e., $P_S(\Obs_S|X)=P_S(\Obs_D|X)$ but the imitator can have different sensors in the target domain.
Optimizing an upper bound of \eqref{eq:imit_prob} that is described in the Supplement gives the following policy to the imitator,
$$
\tilde{\pi}^{(2)}_T(a | \obs_T) \propto \exp\left(\SumInt_{\obs_S}p(\obs_S | \obs_T)\log p_S(a|\obs_S)\right).
$$

\begin{proposition}\label{pro:imi6}
The proposed policy in this case will lead to the following upper bound for  \eqref{eq:imit_prob}, 
$$
\sum_{a, \obs_T, \obs_S} p(\obs_S| \obs_T) p_T(\obs_T)\tilde{\pi}^{(2)}_T(a|\obs_T)\log \frac{\tilde{\pi}^{(2)}_T(a|\obs_T)}{p_S(a|\obs_S)}.
$$
\end{proposition}

Note that in an extreme setting when $\Obs_S$ is determined uniquely from $\Obs_T$, it is straightforward to show that the upper bound in Proposition \ref{pro:imi6} becomes zero. Thus, the proposed proxy leads to the demonstrator's behavior.


\paragraph{Third case:}
 This is the general case where all sensors can be different.
Note that Example 3 belongs to this case. 
 Here, we propose the following policy for the imitator
$$
\tilde{\pi}^{(3)}_T(a|\obs_T) \propto \exp\left(\SumInt_{x}p(x|\obs_T)\log \tilde{p}(a|x)\right), 
$$
where 
$$
\tilde{p}(a|x):=\SumInt_\obs p_S(a|\Obs_S=y)p_S(\Obs_S=y|x).
$$
We introduced the other two cases since they occur frequently in different applications and we can derive theoretical bounds for them.

\section{Experiments}
\label{sec:exp}

In this section we perform experiments for some of the methods proposed in Sections \ref{sec:general}, \ref{sec:dea} and \ref{sec:imitask}.

\begin{figure*}
	\centering
	\includegraphics[height=5.8cm]{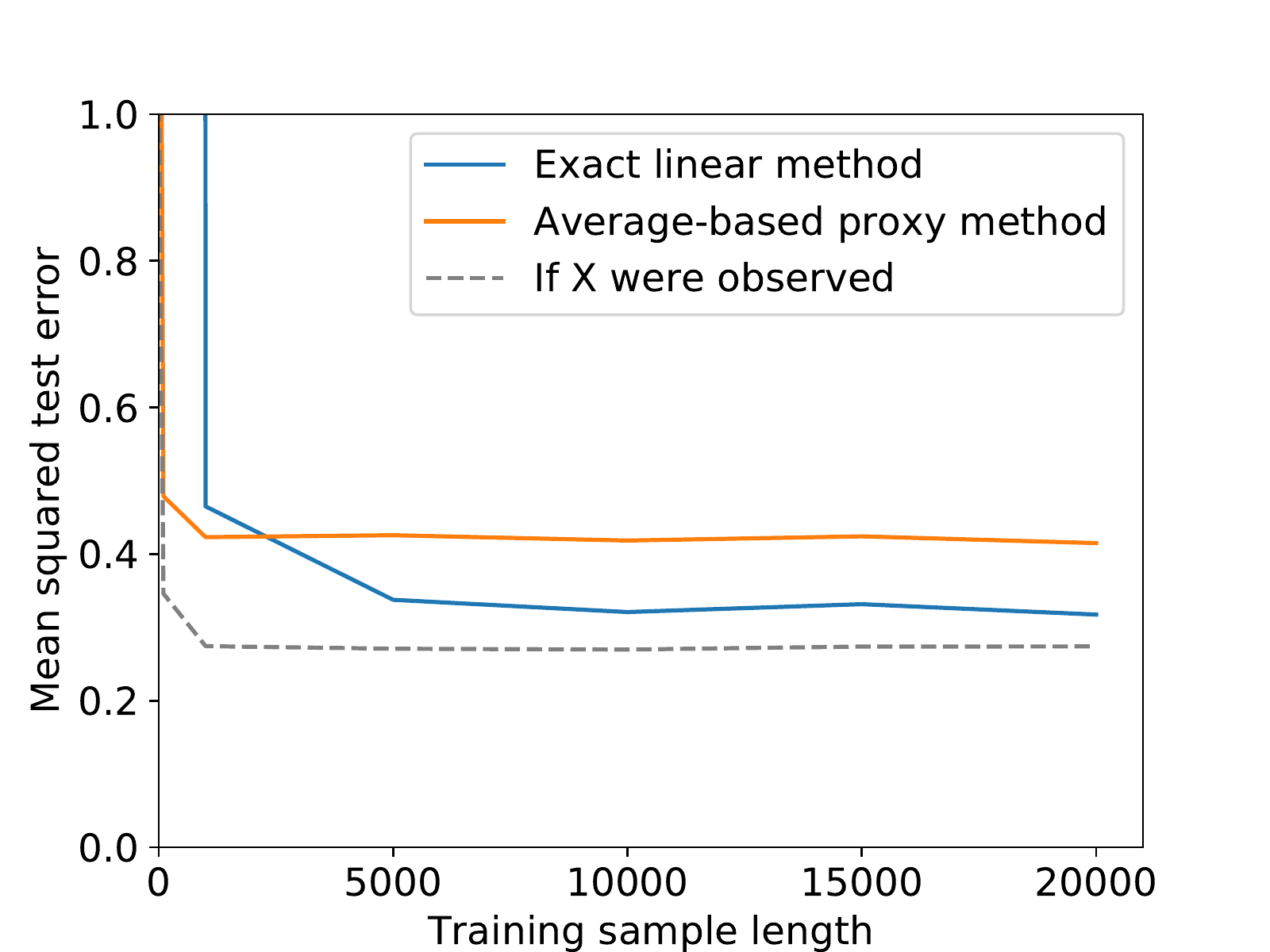} 
	\hspace{-.4cm}
	\includegraphics[height=5.5cm]{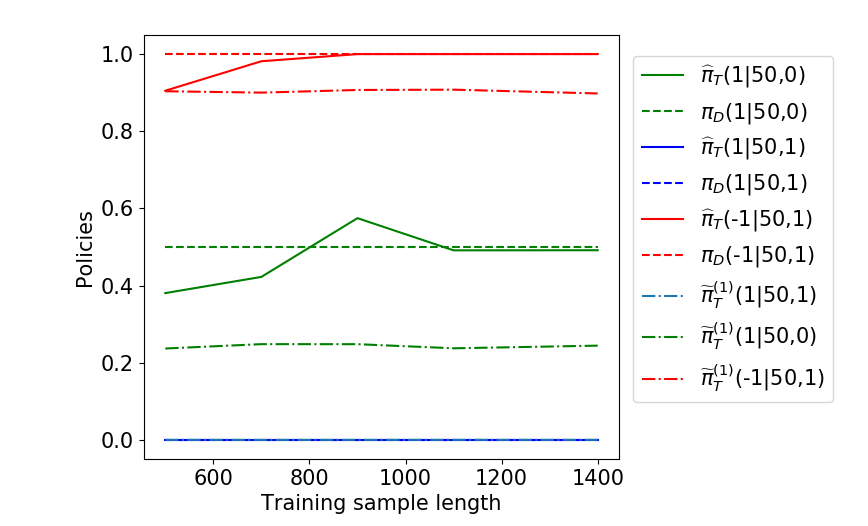}
	\caption{\textbf{Left:} Outcome for the action-effect learning experiment. Our exact linear transfer method (Algorithm \ref{alg:lin}) has higher \emph{variance}, but outperforms the average-based proxy method (sample-level version of \eqref{eqn:avp} for linear case), which can be seen as a baseline, for \emph{long enough samples}. We also plot what could be achieved if $X$ was fully observed in the source domain, as a lower bound. \textbf{Right:} Learned policies for the imitation learning experiment: the true policy $\pi_D$, the policy from the method in Section \ref{sec:imex}, $\hat{\pi}_T$, and the corresponding proxy $\tilde{\pi}^{(1)}_T$.
		The three policies are evaluated at three different points $(a|V_o, b_o) \in \{(1|50,0), (1|50,1), (-1|50,1)\}$.}
	\label{fig:explin}
\end{figure*}

\subsection{{Action-effect learning task}}

\paragraph{Setup:}
In this experiment, we test two of our methods for the \dec: Algorithm \ref{alg:lin} and the proxy in \eqref{eqn:avp} (more specifically: a sample-level version of it for the linear case).
We use the real-world data set ``highD'' \cite{highDdataset} that consists of recordings by drones that flew over several highway sections in Germany (mentioned in Example \ref{expl:i}).
From this data set, we selected all situations, where there is a lead car -- the demonstrator (this is a different setup than Example \ref{expl:i}\footnote{While this is the data set mentioned in Example 1, here we do not consider the indicator lights, since for them we would not have the ground truth.}) -- and a following car on the same lane (which are less than 50m from each other, and have speed at least 80km/h).
Here $X$ is distance, velocities, and acceleration of the follower; $A$ is the acceleration of the demonstrator; and $Z$ is the acceleration of the follower, 1.5 seconds later.

Furthermore, the source domain's $Y_S$ is generated by a randomly drawn matrix $F$ applied to $X$ plus Gaussian noise (as in \eqref{eqn:c1}).
This semi-real approach allows us to have ground truth samples from $P(Z, A, X) = P_T(Z, A, Y_T)$, i.e., the target domain (recall our Assumption \ref{asm:d}).
We apply the two methods on training samples from the source domain $P_S(Z, A, Y_S)$ up to length 20000, and calculate the means (over 20 different data and synthetic noise samples) squared error on separate test samples of length 1000 from $P(Z, A, X)$.

\paragraph{Outcome:}
The outcome for this experiment is \emph{depicted and discussed} in Figure \ref{fig:explin}.

\subsection{Imitation learning task}\label{sec:exp_imi}


\paragraph{Setup:}
In this experiment we simulated the driving scene illustrated in Figure \ref{fig:example1}. The observation set of the demonstrator $\Obs_D$ contains the speed $v_o\in\{40, 45, ...,60\}$ km/h and the indicator light $b_o\in\{0,1\}$ of the lead vehicle. The imitator only gets to see a noisy observation of the demonstrator's speed, i.e., $\Obs_S= v_d + N$, where $N\sim\mathcal{N}(0,1/4)$.  Actions are $-1, +1, 0$ denoting speed reduction by 5km/h, increasing it by 5km/h, and keep the same speed, respectively. In this experiment, we assumed $\Obs_D=\Obs_T$.

We defined the demonstrator's policy to reduce the speed when the indicator of the other vehicle is on $b_o=1$ and increase its speed or keep the same speed when $b_o=0$. 
Note that the classical imitation learning approach will fail in this setting since $\Obs_T\neq\Obs_S$.

We applied Algorithm \ref{alg:svd} plus a criterion to obtain the policy $\tilde{\pi}_T^{(1)}$ for the imitator
This criterion (that is described in the supplement) ensures that the imitator neither increases its speed when $b_o=1$ nor decreases its speed with the same probability when $b_o=0$. We formulated this as a linear programming. 

\paragraph{Outcome:} 
 
Figure \ref{fig:explin} compares the true policy $\pi_D$, the policy from the method in Section \ref{sec:imex}, $\hat{\pi}_T$, and the corresponding proxy $\tilde{\pi}^{(1)}_T$ for different sample sizes.

\section{Conclusions}

Sensor-shift is a significant problem in learning from demonstrations. In this work, we proposed a principled and general framework to address it, based on causal modeling.
We developed novel algorithms that uniquely identify or constrain/approximate the relevant causal effects, and established theoretical guarantees.
The take away message is that the relevant causal relationships may still be identifiable, even if the demonstrator, spectator and target agent have different sensors.



\bibliography{causal_traffic}
\bibliographystyle{aaai}

\end{document}